\documentclass[10pt,twocolumn,letterpaper]{article}

\usepackage[pagenumbers]{iccv} 

\definecolor{iccvblue}{rgb}{0.21,0.49,0.74}
\usepackage[pagebackref,breaklinks,colorlinks,allcolors=iccvblue]{hyperref}
\usepackage{bm}
\usepackage{multirow}

\def\paperID{***} 
\def\confName{ICCV}
\def\confYear{2025}

\title{VideoScan: Enabling Efficient Streaming Video Understanding via Frame-level Semantic Carriers}

\author{
	\textbf{Ruanjun Li}$^{1,2}$ \thanks{The work was done during an internship at TeleAI.}
	\quad
	\textbf{Yuedong Tan}$^{1,3}$
	\quad
	\textbf{Yuanming Shi}$^{2}$
	\quad
	\textbf{Jiawei Shao}$^{1}$ \thanks{Corresponding author}
	\\
        $^{1}$TeleAI, China Telecom
        \quad
	$^{2}$ShanghaiTech University
	\quad 
	$^{3}$Xidian University
}

\begin{document}
\maketitle

\begin{abstract}
	This paper introduces VideoScan, an efficient vision-language model (VLM) inference framework designed for real-time video interaction that effectively comprehends and retains streamed video inputs while delivering rapid and accurate responses. 
	A longstanding challenge in video understanding—particularly for long-term or real-time applications—stems from the substantial computational overhead caused by the extensive length of visual tokens. 
	To address this, VideoScan employs a single semantic carrier token to represent each frame, progressively reducing computational and memory overhead during its two-phase inference process: prefilling and decoding.
	The embedding of the semantic carrier token is derived from an optimized aggregation of frame-level visual features, ensuring compact yet semantically rich representations. 
	Critically, the corresponding key-value pairs are trained to retain contextual semantics from prior frames, enabling efficient memory management without sacrificing temporal coherence. 
	During inference, the visual tokens of each frame are processed only once during the prefilling phase and subsequently discarded in the decoding stage, eliminating redundant computations. 
	This design ensures efficient VLM inference even under stringent real-time constraints.
    Comprehensive experiments on diverse offline and online benchmarks demonstrate that LLaVA-Video, supported by our method, achieves up to $5\times$ and $1.29\times$ speedups compared to its original version and previous efficient streaming video understanding approaches, respectively.
    Crucially, these improvements are attained while maintaining competitive performance and ensuring stable GPU memory consumption (consistently $\sim 18GB$, independent of video duration).
	
\end{abstract} 
\section{Introduction}
\label{sec:intro}

\begin{figure}
	\includegraphics[width=0.98\linewidth]{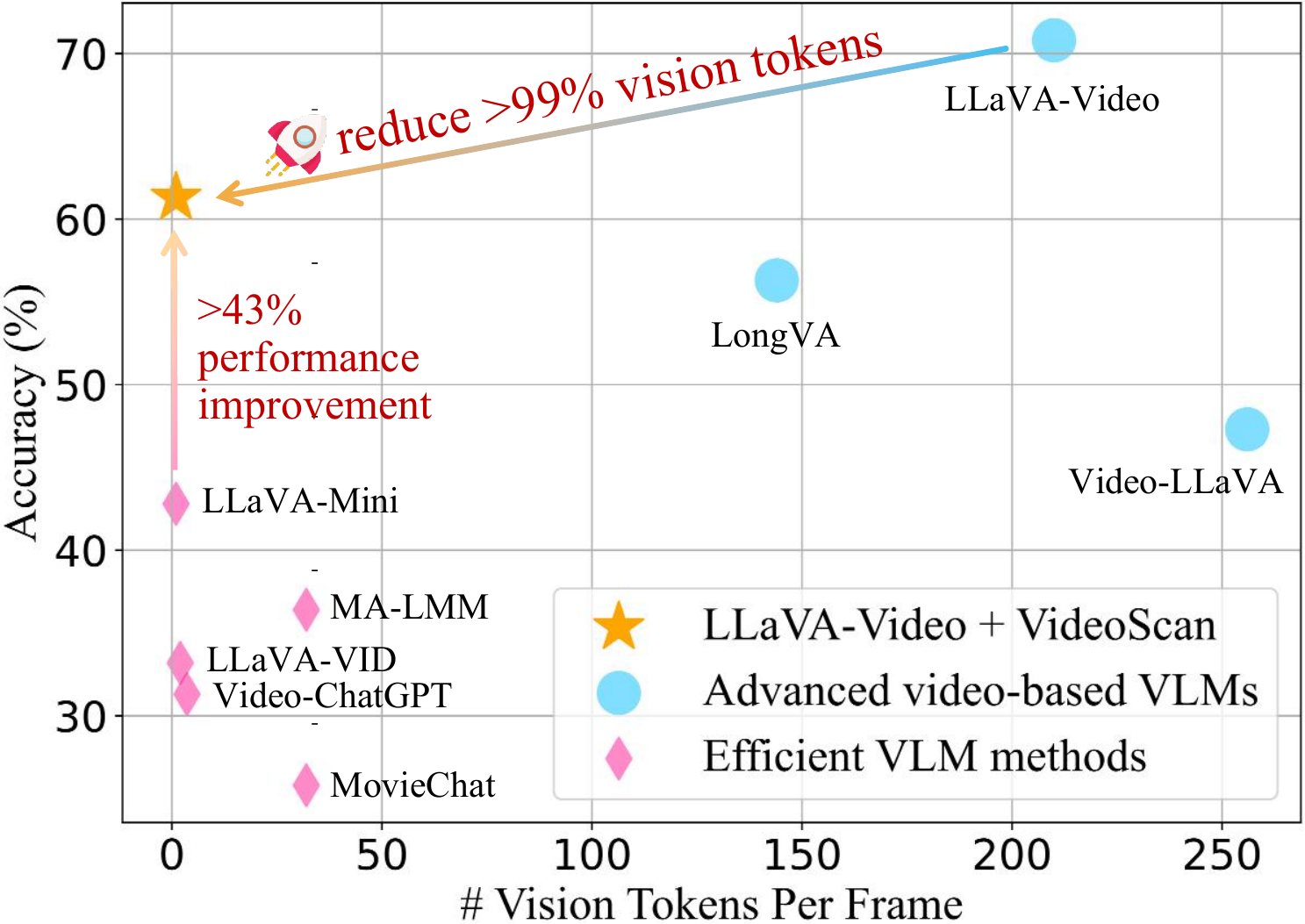}
	\caption{Performance comparison on MLVU benchmark. Supported by the efficiency of VideoScan, LLaVA-Video reduces $>99$\% of vision tokens in inference while maintaining comparable performance, which achieves $>45$\% improvement the performance of LLaVA-Mini with only one vision token for each frame.    
	}
	\vspace{-0.5cm}
	\label{fig:compare}
\end{figure}

Building on the success of large language models (LLMs) \cite{openai2024gpt4o,team2023gemini,bai2023qwen, yang2024qwen2}, researchers have expanded their capabilities to computer vision by developing vision-language models (VLMs) \cite{openai2023gpt4v,radford2021learning,li2024llava,li2022blip,li2023blip,wang2024internvideo2, luo2023valley, wang2024qwen2}. 
These models combine text and visual data to better understand and generate content, such as describing images or answering questions about visual scenes. 
Recent advancements in video-based VLMs \cite{lin2023video, zhang2023video, liu2024st, wang2024internvideo2, zhang2024video, liu2024world} have further extended these capabilities to tasks like video captioning and action recognition by incorporating techniques to process temporal dynamics, align text with video, and efficiently encode visual features.
Thanks to their strong reasoning skills, video-based VLMs now power applications \cite{shao2025ai} such as robotics \cite{wang2024large, mu2023embodiedgpt, khandelwal2022simple}, augmented reality (AR) devices \cite{chen2024videollm}, and interactive video assistants \cite{maaz2023video, li2023videochat}.
However, many of these applications require real-time performance —for example, robots need instant feedback to navigate environments safely \cite{wang2024large}, and AR glasses must process live video streams without delays \cite{mu2023embodiedgpt}.
To meet these demands, streaming video understanding frameworks \cite{chen2024videollm,di2025streaming, qian2025streaming} have been proposed. These systems process videos as continuous streams of frames, enabling immediate analysis and interaction.
Yet, a major challenge persists: the computation overhead of processing massive visual inputs \cite{liu2024world}.
Each video frame generates a large number of visual tokens, which strains hardware resources and causes huge end-to-end response delays, making real-time performance difficult to achieve.

Among the recent works on video understanding with LLMs \cite{lin2023video,jin2024chat,zhang2024long, liu2024st, ren2024timechat}, a prevalent strategy is using sparse sampling or pooling \cite{bolyatoken,shang2024llava, li2024tokenpacker}, in frame, patch or even token level, to achieve $30\% \sim 50\%$ vision token reduction rate.
However, in video-based scenarios, this is insufficient for long-term video interactions with over hundreds of frames.
To address this limitation, a natural idea \cite{song2024moviechat,li2024llama,zhang2025llava} is to develop frame-level compression techniques that leverage strategically selected or synthesized tokens embedded with high-density visual features. 
Some works \cite{ryoo2021tokenlearner, li2023blip} introduce learnable modules to query the visual features and extract general vectors for frames.
However, excluding the hard to learn manner, it is also difficult to maintain acceptable performance under high compression rate with tokens extracted from simple modules.
Another approaches \cite{li2024llama, zhang2025llava} explore instruction-guided vision token compression or fusion methods to achieve an extreme compression rate with $2$ or even $1$ token for each frame while maintaining an competitive performance. 
As for streaming video understanding, an extra storage module, namely memory bank, is commonly employed to store historical visual features \cite{zhang2024flash, qian2025streaming, di2025streaming}. 
It is a promising way to support real-time video interaction with retrievable visual information.
To this end, the query-dependent compression pipeline limits the ability to produce a general representation that can serve as information backups and handle diverse and temporal instructions in real-time video interactions.

With the improved exploration about the inner pattern of transformer decoder architecture, a pouch of works \cite{chen2024image, zhang2024cls} explore vision token reduction within backbone LLMs through interpretable analysis in attention mechanism.
A typical work is FastV \cite{chen2024image}, which dives into the attention mechanism and employs an importance-based manner to reduce vision token.
Although they do not straightly reduce the input token number before LLM, and also can not serve for huge reduction rate in video for the lack of appropriate training strategy, these works really provide great insights of the inner pattern of LLMs.

Based on these insights, in this work, we take a step deeper in the attention mechanism and the transformer architecture, and leverage the associated characteristics to support extreme vision token reduction.
We propose VideoScan, a novel and efficient inference framework for streaming video understanding that represents each frame by only one single token. 
The key innovation lies in the construction of a specialized \emph{semantic carrier token}, which leverages the in-context summarization capabilities of LLMs to encapsulate all frame-level visual information into a compact representation.
 
The main contributions of this work are threefold:
 \begin{enumerate}[1)]
    \item \textbf{Semantic Flow Compression:} We introduce a novel semantic carrier token that leverages the inherent semantic flow in backbone LLMs to reduce visual token complexity. A two-stage training strategy is proposed with semantic-aware causal masking progressively to enhance the semantic flow.
    \item \textbf{Real-time Streaming Video Framework:} An efficient VLM framework, namely VideoScan, is proposed to support real-time streaming video understanding with one semantic carrier token per frame. We further employ a memory mechanism to maintain information-rich context while ensuring stable and optimized GPU memory usage (e.g., $18$GB) independent of video duration.
    \item \textbf{Balance Between Effectiveness and Efficiency:} Extensive experiments demonstrate that VideoScan, with one token per frame, achieves real-time inference with 6 serving FPS on consumer GPUs (e.g., RTX 4090) -- $\sim 5 \times$ acceleration over LLaVA-Video -- while maintains competitive accuracy, striking a balance between computational efficiency and task effectiveness.
\end{enumerate}

\begin{figure*}
	\includegraphics[width=0.98\linewidth]{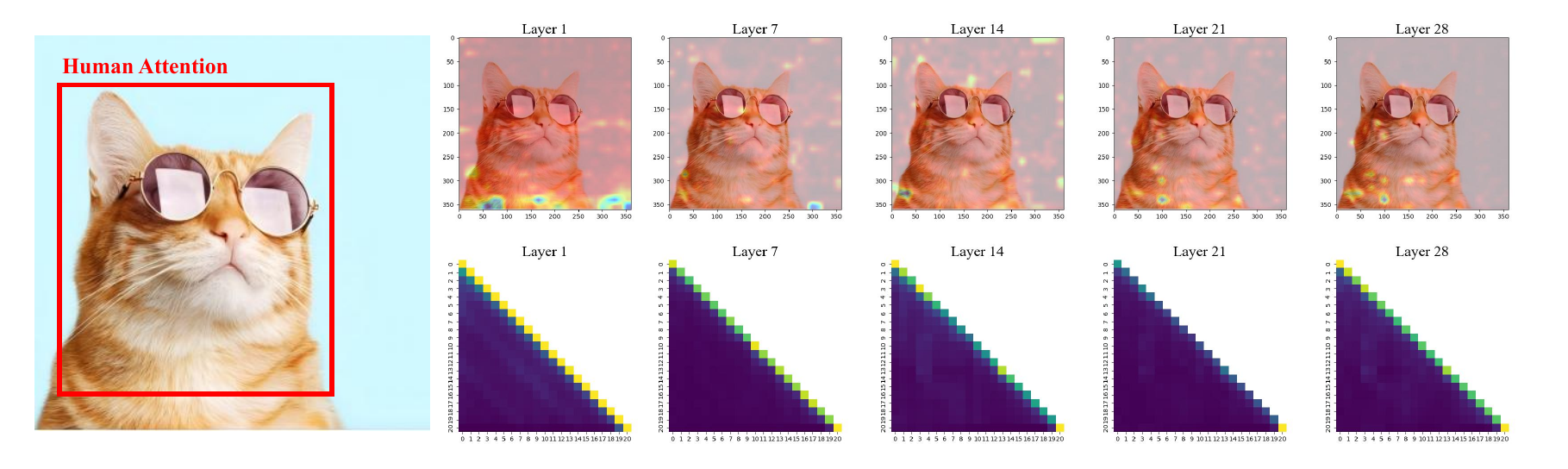}
	\caption{An example of attention map. For better visualization, we truncate the tokens corresponding to system instructions. The attention map reveals an 'attention sink,' where the model tends to assign higher scores to tokens with nearby positions. Meanwhile, when projecting the averaged attention score at generated tokens onto the original image, it becomes evident that the visual tokens the model focuses on differ from those humans perceive as important, and somehow they tend to be near the response. 
	}
	\vspace{-0.5cm}
	\label{fig: short}
\end{figure*}

\section{Related Work}
\label{sec:related}


\noindent\textbf{Video-based Vision Language Model.}
Building upon advancements in image-text understanding  \cite{radford2021learning, li2022blip, li2023blip, liu2024visual}, recent efforts have extended VLMs to the video domain.
For instance, Video-ChatGPT \cite{maaz2023video} pioneers this direction by adapting the LLaVA \cite{liu2024visual} framework with a temporal-aware visual encoder, enabling VLMs to understand video.
Later advancements prioritize visual and textual contexts aligning \cite{lin2023video, liu2024st, zhang2023video, wang2024internvideo2, zhang2024video}, fine-grained feature capturing \cite{jin2024chat}, and enhanced spatiotemporal reasoning \cite{zhang2023video, ren2024timechat}.
Recently, an increasing number of studies \cite{li2023videochat,song2024moviechat, ren2024timechat, wang2024videoagent, he2024ma, wang2024internvideo2} have explored methods to construct efficient VLMs by compressing visual inputs while enhancing spatial-temporal cues, thereby supporting long-term video understanding.

\noindent\textbf{Streaming Video Understanding.}
Real-time video-based VLM inference plays a crucial role in enabling various downstream tasks, such as embodied AI \cite{mu2023embodiedgpt, khandelwal2022simple}, temporal action recognition \cite{xing2023svformer}, and video dialogue \cite{li2023videochat, maaz2023video, song2024moviechat}.
Unlike conventional offline processing, streaming video understanding treats inputs as continuous frame sequences, allowing VLMs to interact with live visual data for instantaneous decision-making.
VideoLLM-Online \cite{chen2024videollm} is a leading work that discussed the inference framework of video streaming dialogue and allowed the VLM to respond at a proper point.
Subsequent efforts \cite{zhang2024flash, qian2025streaming, di2025streaming} further optimize temporal modeling through memory-argument architectures, enhancing the storage and retrieval of spatiotemporal cues via compressed token buffers and adaptive attention mechanisms.
A persistent challenge lies in the inherent computational asymmetry between high-dimensional visual tokens and lightweight linguistic representations, which complicates efficient real-time interaction.
Current methods often struggle to balance compressed vision token length, memory overhead, and accuracy for long-duration streams.
To address this gap, our work focuses on computation- and memory-efficient streaming architectures, achieving significant acceleration through converting all frame-level information to one token.

\noindent\textbf{Vision Token Reduction.}
Token reduction in VLMs has emerged as a promising approach to streamlining computational demands by shortening input token sequences, thereby enabling more efficient inference.  
For video-based VLMs, this strategy tackles a critical challenge: reducing the number of visual tokens per frame allows models to process longer video sequences within fixed input length constraints, a key requirement for robust long-form video understanding and real-time video dialogue.
Representative techniques \cite{yao2024deco} include Token Merging \cite{bolyatoken}, PruMerge \cite{shang2024llava}, and TokenPacker \cite{li2024tokenpacker}, by merging or pruning spatially or temporally similar tokens. 
Other attempts \cite{bai2023qwen, ryoo2021tokenlearner}, leverage token generation modules, like a Q-Former \cite{li2023blip}, to project variable-length visual tokens into compact representations.
These studies rely solely on visual inputs and struggle to maintain accuracy at high compression rates.
Another line of work \cite{zhang2024sparsevlm, li2024llama, zhang2025llava} achieves extreme compression of visual inputs to just 2 or 1 token by leveraging task-specific instructions to filter out task-irrelevant information or generate global representations.
Though these methods achieve state-of-the-art compression rates without significant performance degradation, their instruction-dependent designs tightly couple visual information with query-specific context, hindering their applicability in streaming video understanding scenarios where reusable memory and temporal coherence are essential.
To address these limitations, we propose a token-efficient framework for streaming video inference. 
Our approach represents each frame with a single token while avoiding instruction-specific context leakage, enabling fast video interaction, and preserving reusable memory for long-term temporal reasoning.

\begin{figure*}[htbp]
	\centering
	\includegraphics[width=0.98\linewidth]{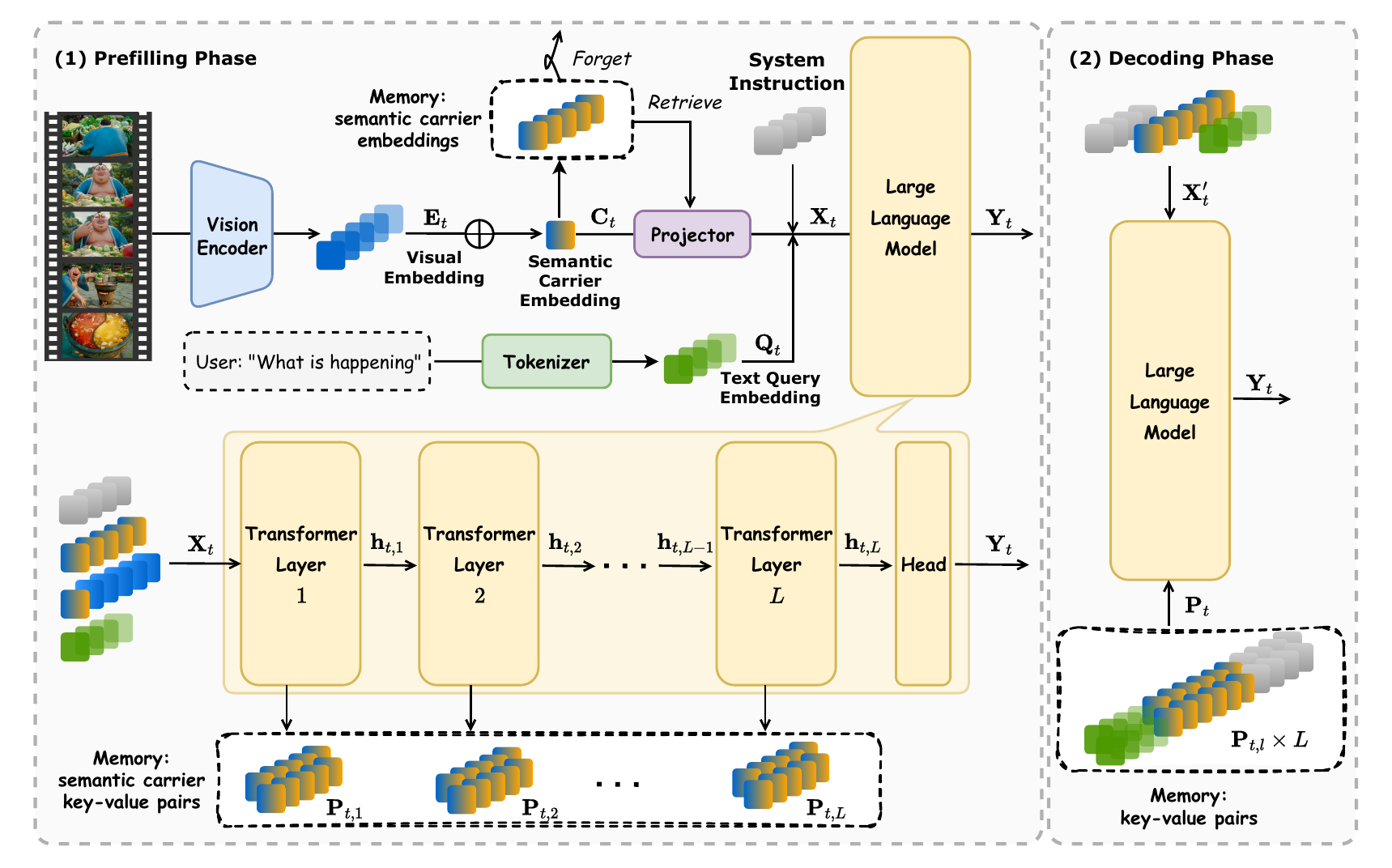}
	\caption{The overall workflow of the proposed VideoScan inference framework. We construct a semantic carrier token by \textbf{the frame level visual features through an average pooling}, and leverage it to \textbf{inherit all in-context semantic information in KV}. Frame-level visual tokens are processed exclusively during the prefilling phase and subsequently discarded. All visual information required during decoding is sourced from the semantic carrier. We also introduce \textbf{a memory mechanism}, which stores the embeddings and KV of semantic carriers, to support long-term video interactions with retrievable past visual information and optimized GPU memory usage.}
	\label{fig: inference framework}
	\vspace{-0.5cm}
\end{figure*}

\noindent\textbf{Semantic Flow.}
In decoder-only LLMs, an attention window \cite{xiao2023efficient, fan2023parallel, xiaoefficient, xiao2024duoattention} usually observes that tokens with close relative positions will receive more attention.
Inspired by such phenomenon, recent research \cite{chen2024image, hu2024illava, zhang2025llava, wen2025stop, xing2024pyramiddrop} has further explored the semantic flow within this architecture.
A notable work is FastV \cite{chen2024image}, which provided an in-depth analysis of the attention maps of different layers, showing that information carried by the input token will be obtained at shallow layers.
They further claimed that vision tokens play a significant role only in the early layers of the backbone model and cut down the less important parts in deeper layers for efficacy.
Our work shares similar observations with them and steps deeper to utilize and enhance the characteristics to support efficient video-based VLM inference in stream.

\section{Methods}
\label{sec:methods}

In this section, we first introduce semantic carrier, a natural semantic summation and compression in the frame level.
As shown in \cref{fig: inference framework}, we propose a video-based VLM inference acceleration framework, namely VideoScan, which supports efficient streaming video understanding with extended video length, minimized computation latency, and optimized memory cost.

\subsection{Semantic Flow}
\label{sec:semantic flow}
We begin by introducing our observation in the decoder transformer architecture.
As shown in \cref{fig: short}, the attention maps across different transformer layers exhibit distinct distributional characteristics. 
In shallow layers, the model allocates attention relatively uniformly across all tokens, whereas deeper layers demonstrate a positional bias, prioritizing tokens with relative proximity and those system instructions located at preceding absolute positions.
Recent works \cite{chen2024image} demonstrated that removing up to $90\%$ of vision tokens with lower attention scores after shallow layer does not lead to performance degradation.
This finding inspired us to explore how the information is transferred in LLMs.
We first dive into the mathematical representation of the self-attention mechanism. 
Given input embeddings $\bm{X}=[\bm{x}_1, \bm{x}_2, \cdots, \bm{x}_s]^T$, where $\bm{X} \in \mathbb{R}^{s\times d}$, $s$ is the sequence length and $d$ is the embedding dimension, the output can be formulated as:
\begin{equation}
	\bm{Y} =  \text{Softmax}\left( \frac{\bm{Q}\bm{K}^T}{\sqrt{d_k}} \right)\bm{V} .
\end{equation}
Here $\bm{Q}=\bm{X}\bm{W}_{Q}$, $\bm{K}=\bm{X}\bm{W}_{K}$, $\bm{V}=\bm{X}\bm{W}_{V} \in \mathbb{R}^{d,d_k}$ are learnable linear projectors.
When inferring the $(s+1)$-th token, the output $Y$ is derived as a weighted aggregation of the value matrix $V$. 
This implies that the representation of the $(s+1)$-th token at later layers inherently encapsulates a summarization of the preceding $s$ tokens through $\bm{Y}$. 
Notably, such semantic aggregation is propagated sequentially through cascaded transformer layers, facilitating progressive refinement and contextual coherence across the sequence.
This conclusion is somewhat consistent with the experimental findings reported in \cite{chen2024image}, which demonstrates that when input visual tokens with low attention scores are pruned after the $l$-th transformer layer, the performance degradation diminishes as $l$ increases. 
The difference is that LLMs do not prefer tokens precisely where visual information is revealed as we human like.
Instead, they can effectively derive knowledge from tokens following those with high attention scores.

To further validate this hypothesis, we conducted visualization experiments using the LLaVA-OneVision \cite{li2024llava} model on single-image inference tasks. 
We present the visualization results of one sample in \cref{fig: short}. 
The heatmap is the averaged attention map of all generated tokens, showing that the model does not focus on those input visual tokens corresponding to the subject's location (in this case, the cat).  
Other experimental findings reveal that even when tokens directly corresponding to the primary subject of the image are removed, LLM is still capable of completing the inference task by leveraging the last few tokens with semantically aggregated information from shallower layers. 
The observation, called semantic flow, underscores the progressive nature of information aggregation and refinement across deeper layers of the Transformer architecture.

\begin{figure}[t]
	\centering
	\includegraphics[width=0.98\linewidth]{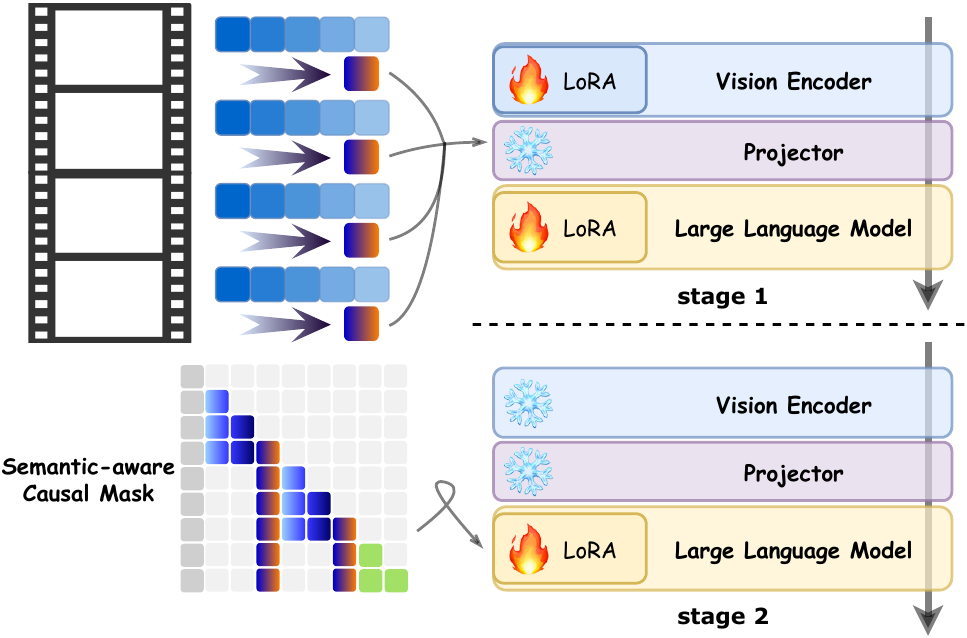}
	\caption{The proposed two-stage training recipe for VideoScan. At stage 1, each frame is represented by a semantic carrier token. The visual inputs are semantic carrier tokens only. At stage 2, the semantic token is positioned at the end of each frame. A semantic-aware causal mask is implemented to enhance the semantic flow in KV, maintaining that the LLM only accesses the semantic carrier.
	}
	\vspace{-0.5cm}
	\label{fig: training recipe}
\end{figure}

\subsection{Semantic carrier}
Inspired by the natural information aggregation ability of semantic flow, in this paper, we introduce a frame-level vision information distillation, especially for real-time video interaction scenarios.
The design concept is highly intuitive with a specialized token utilized to encapsulate the semantic information aggregated from the preceding context. 
This token, namely the semantic carrier token, can be decomposed into two key components: (1) an input embedding that consolidates frame-level information, and (2) a semantic flow inherited through reusable KV. 
The input embedding comes from an average pooling of current frame token embeddings and serves as an aggregated and smoothed representation of the frame-level visual features, enabling a more stable and concise summarization of the visual content.
Furthermore, since the input embedding is derived from the vision encoder, it is straightforward to learn and adapt.
To leverage the inherent summarization capability of semantic flow, we introduce a KV-based semantic aggregation mechanism tailored for streaming video scenarios.
As illustrated in \cref{sec:semantic flow}, visual information is propagated between layers through the sequential transmission of token representations.
Therefore, the semantic carrier is positioned at the end of the vision tokens corresponding to the current frame. 
During inference, the semantic carrier stores the associated KV in the cache memory, enabling it to accumulate and propagate contextual information in different transformer layers within and across frames.
We further implement a well-designed training strategy to enhance the semantic flow, which will be detailed in \cref{sec:training strategy}.

\subsection{VideoScan Inference Framework}
Based on the semantic carrier, the overall inference workflow of the proposed VideoScan framework is illustrated in \cref{fig: inference framework}, comprising two phases: prefilling and decoding.
For real-time video interaction, we model visual inputs as a continuous frame-by-frame stream. 
Assuming a frame rate of $1$ frame per second, at time step $t$, the current frame $V^t \in \mathbb{R}^{H \times W \times 3}$ is processed as follows:

\noindent\textbf{Prefilling Phase: }
$V^t$ is encoded via a transformer-based vision encoder \cite{radford2021learning}, yielding visual embeddings $\bm{E}^t \in \mathbb{R}^{N \times d}$, where $N$ is the number of visual tokens per frame and $d$ is the embedding dimension.
A semantic carrier token $\bm{C}^t$ is constructed by average pooling over $\bm{E}^t$: $\bm{C}^t := \frac{1}{N} \sum_{i=1}^N \bm{E}^t, \quad \bm{C}^t \in \mathbb{R}^{1 \times d}$.
$\bm{C}_t$ captures global frame-level semantics and is stored in a memory bank $\mathcal{M}$.
Up to $M$ historical semantic carrier tokens are retrieved from $\mathcal{M}$.
After the projector alignment module, we concatenate all input information (i.e., the system instruction, visual, and textual tokens) and get the input of backbone LLM for the first forward and generate the first token.

\noindent\textbf{Decoding Phase: }
Frame-level visual tokens undergo processing exclusively during the initial prefilling stage and are discarded post-processing.
All visual information required for decoding is retrieved from the semantic carrier tokens, eliminating redundant computations and minimizing memory consumption.

\noindent\textbf{Memory Mechanism:}
To advance long-term streaming video understanding, we propose a memory mechanism that dynamically retains salient historical context through selective preservation.
The memory bank $\mathcal{M}$ maintains semantic carrier tokens with max length of $M$, storing their embeddings and associated KV pairs, ensuring stable GPU memory consumption while optimizing computational efficiency.
When capacity is reached, a feature duplication-based eviction strategy is employed to manage information retention.
Specifically, the mechanism compares embeddings of adjacent semantic carriers, discarding the older carrier exhibiting the largest cosine similarity to the incoming token.

All these designs guarantee a compact yet semantically rich visual representation, delivering robust temporal reasoning, consistent computational and memory efficiency, and great scalability even during prolonged inference on extensive video streams.

\subsection{Training strategies}
\label{sec:training strategy}
Since the VideoScan takes a progressive information compacting paradigm, structuring its processing pipeline into two distinct phases: prefilling and decoding,
we introduce a two-stage training strategy to align with the architecture and progressively enhance the ability of the semantic carrier token to summarize both global representation embeddings and KV pairs, as shown in \cref{fig: training recipe}.
In the first stage, the vision encoder and LLM backbone are jointly fine-tuned using low-rank adapters (LoRA).
This stage focuses on adapting the vision encoder to align with the average-pooled semantic carrier representation and adjusting the LLM backbone to handle single-token-per-frame inputs while preserving contextual understanding.
We adopt a semantic flow-aware training as stage two with only LLM fine-tuned.
To strengthen temporal coherence, we introduce a semantic-aware causal mask seen in \cref{fig: training recipe}. 
This mask enforces an unidirectional semantic flow.
The visual information of each frame is captured by its semantic carrier token, and subsequent tokens (e.g., text or future frame carriers) must rely on prior semantic carriers for prediction, preventing information leakage to future frames.

\begin{table*}[]
	\centering
	\caption{Performance on offline video question-answering benchmarks. `\#Frames' is the number of sampled frames. `\#Tokens' represents the number of vision tokens for each frame. `MVB' refers to MVBench. `LVB' refers to LongVideoBench. `V-MME' refers to VideoMME benchmark. All the listed scores are in percentage (\%).}
	\label{tab: offline}
	{
		\small
		\begin{tabular}{lcccccccccc}
			\toprule
			\multirow{2}*{\textbf{Method}}& \multirow{2}*{\textbf{\#Frames}}& \multirow{2}*{\textbf{\#Tokens}}& \multirow{2}*{\textbf{MVB}}& \multirow{2}*{\textbf{MLVU}}&\multirow{2}*{\textbf{LVB}}& \multicolumn{4}{c}{\textbf{V-MME (w/o. sub)}} & \textbf{Avg.} \\ 
			\cmidrule(lr){7-10}
			~&~&~&~&~&~ & \textbf{Overall} & \textbf{Short} & \textbf{Medium} & \textbf{Long} & (\%) \\ 
			\midrule
			\multicolumn{4}{l}{\textit{Close-source advanced video-based VLMs}}\\
			\midrule
			GPT-4V \cite{openai2023gpt4v} & - & - & -& 49.2 & 61.3 & 59.9 &70.5&55.8&53.5&-\\
			GPT-4o \cite{openai2024gpt4o} &  384 & - &- & 64.6 & 66.7 & 71.9 &80.0&71.3&62.2&-\\
			Gemini-1.5-Flash \cite{team2023gemini} &  1/0.5fps & - & -& -& 61.6 & 70.3&78.8&68.8&61.1&-\\
			Gemini-1.5-Pro \cite{team2023gemini} &  1/0.5fps & - & -&- & 64.0 & 75.0&81.7&74.3&67.4&-\\
			\midrule
			\multicolumn{4}{l}{\textit{Open-source advanced video-based VLMs (7B)}}\\
			\midrule
			Video-LLaMA \cite{zhang2023video}  & 16&64 & 34.1&48.5&-&-&-&- &- &-\\
			Video-LLaVA \cite{lin2023video}  & 8 & 256 & 43.5 & 47.3 &37.6 &39.9&45.3&38.0&36.2&- \\
			LongVA \cite{zhang2024long}  &128&144&-&56.3&-&52.6&61.1&50.4&46.2&-\\
			LLaVA-OV-7B \cite{li2024llava}& 64 & 196& \underline{56.7}&\underline{64.7}&\underline{56.3}&\underline{58.2}&\underline{70.2}&\underline{56.6}&\underline{47.7}&\underline{59.0}\\
			LLaVA-Video-7B \cite{zhang2024video}  & 64 & 210 & \textbf{58.6}  & \textbf{70.8} & \textbf{58.2 }& \textbf{63.3}&\textbf{75.4}&\textbf{62.6}&\textbf{51.8}&\textbf{62.7} \\
			\midrule
			\multicolumn{4}{l}{\textit{Our proposed VideoScan based on LLaVA-Video-7B}} \\
			\midrule
			\textbf{VideoScan ($M=64$)}  & \textbf{1fps} & \textbf{1} &48.9& 59.7 & 47.1 &54.0 & 62.1 & 53.2 & 46.6&52.4\\ 
			\textbf{VideoScan ($M=128$)}  & \textbf{1fps} & \textbf{1} & 48.9 & 61.3 & 49.5 & 55.1& 64.2 & 54.3 & 46.7&53.7\\ 
			\bottomrule 
		\end{tabular}
	}
\end{table*}

\begin{table*}[]
	\centering
	\caption{Results of comparison with token-efficient methods on MLVU benchmark (\%). Detailed evaluation includes Topic Reasoning (TR), Anomaly Recognition (AR), Needle QA (NQA), Ego Reasoning (ER), Plot QA (PQA), Action Order (AO), and Action Count (AC).}
	\label{tab: offline mlvu}
	\begin{tabular}{lcccccccccc}
		\toprule
		\multirow{2}{*}{\textbf{Method}} & \multirow{2}{*}{\textbf{\#Frames}} & \multirow{2}{*}{\textbf{\#Tokens}}& \multicolumn{2}{c}{\textbf{Holistic}} & \multicolumn{3}{c}{\textbf{Single Detail}} & \multicolumn{2}{c}{\textbf{Multi Detail}} & \multirow{2}{*}{\textbf{Avg.}} \\
		\cmidrule(lr){4-5} \cmidrule(lr){6-8} \cmidrule(lr){9-10}
		~ & ~ &  ~ & \textbf{TR.} & \textbf{AR.} & \textbf{NQA.} & \textbf{ER.} &\textbf{PQA.} & \textbf{AO.}& \textbf{AC.}\\
		\midrule
		MovieChat \cite{song2024moviechat}& 2048&32 &29.5 &25.0 &24.2 &24.7 &25.8 &28.6 &22.8 &25.8\\
		Time-Chat \cite{ren2024timechat} &96& 32&23.1 &27.0 &24.5 &28.4 &25.8 &24.7 &\textbf{32.0} &30.9\\
		MA-LMM \cite{he2024ma} &1000 &32 &51.9 &35.5 &43.1 &38.9 &35.8 &\underline{25.1} &24.3 &36.4\\
		Video-ChatGPT \cite{maaz2023video} &100 &$\sim 3.6$&26.9 &24.0 &40.3 &\underline{42.0} &29.9 &\underline{25.1} &31.1 &31.3\\
		LLaMA-VID \cite{li2024llama} & 1fps & 2 & 50.8&34.5&30.1&32.7&32.5&23.9&27.8&33.2 \\
		LLaVA-Mini \cite{zhang2025llava}  & 1fps & 1 & \underline{76.0}&\underline{50.0}&\underline{44.5}&37.5&\underline{49.0}&24.3&18.4&\underline{42.8} \\
		\midrule
		\textbf{VideoScan ($M=128$)} & 1fps & 1 & \textbf{81.8}&\textbf{64.0}&\textbf{67.9}&\textbf{62.5}&\textbf{64.9}&\textbf{44.0}&\underline{31.6}&\textbf{61.3} \\ 
		\bottomrule
	\end{tabular}
\end{table*}

\begin{table*}[]
	\centering
	\caption{Performance on online streaming video question-answering benchmark VStream-QA. The latency and VRAM usage are measured from a single query-to-answer process with the same visual input executed in a single machine with one A100 GPU.}
	\label{tab: online results}
	\begin{tabular}{lccccccccc}
		\toprule
		\multirow{2}{*}{\textbf{Method}} & \multirow{2}{*}{\textbf{FPS}} & \multirow{2}{*}{\textbf{\#Tokens}} & \multicolumn{2}{c}{\textbf{RVS-Ego}} & \multicolumn{2}{c}{\textbf{RVS-Movie}} & \multirow{2}{*}{\textbf{Latency}} & \multirow{2}{*}{\textbf{VRAM}} \\
		\cmidrule(lr){4-5} \cmidrule(lr){6-7}
		~ & ~ &~ & \textbf{Acc.} & \textbf{Sco.} & \textbf{Acc.} & \textbf{Sco.} & & & \\
		\midrule
		MovieChat \cite{song2024moviechat}&2048 &32& 50.7&3.4&36.0&2.3&-&-\\
		LLaVA-VID \cite{li2024llama}&1fps&2&53.4&\underline{3.9}&48.6&3.3&-&-\\
		Flash-VStream-7B \cite{zhang2024flash} & 1fps &-& 57.3 & \textbf{4.0} & 53.1 &3.3&\textbf{2.1s}&\underline{19GB} \\
		ReKV (LLaVA-OV-7B) \cite{di2025streaming}& 1fps & offload & \textbf{63.7} & \textbf{4.0} & \textbf{54.4} &\textbf{3.6} & \underline{2.7s} &36GB \\
		\midrule
		\textbf{VideoScan (M=128)} & 1fps &1& \underline{60.9} & \textbf{4.0} & \underline{54.1} &\underline{ 3.5} &\textbf{2.1s} & \textbf{18GB}\\ 
		\bottomrule
	\end{tabular}
\end{table*}

\section{Experiments}
\label{sec:experiments}
In this section, we present the experimental setup and compare VideoScan with state-of-the-art video-based VLMs on offline video understanding benchmarks and online streaming video question-answering benchmarks, focusing on both its effectiveness and efficiency.
Besides, we conduct an ablation study to evaluate the contributions of each component of VideoScan. 

\subsection{Implementation Details}

The proposed VideoScan inference framework is built upon the LLaVA-Video 7B model \cite{zhang2024video}, which adopts a two-stage training strategy to enhance the semantic carrier token. 
Both stages utilize data sampled from the LLaVA-Video-178K dataset \cite{zhang2024video}, which is the original training data for LLaVA-Video. 
For the first stage, we randomly sampled videos from each subset from LLaVA-Video, stratified by the video duration, to ensure diverse video coverage.
This stage focuses on adapting the vision encoder and LLM backbone to process single-token frame representations.
To enhance dense temporal reasoning, we prioritize videos with rich temporal dynamics in stage two, the semantic-aware training stage.
In inference, VideoScan gets input frames at $1$fps.
Upon reaching the memory bound $M$, which we set 64 and 128 in evaluation, the memory mechanism will selectively forget those depending on the given strategy to maintain a stable memory usage.

\noindent\textbf{Baselines.}
VideoScan strikes a balance between effectiveness and efficiency in video-based vision-language model (VLM) inference. 
To thoroughly evaluate its performance, we benchmark it against state-of-the-art VLMs, including the GPT series \cite{openai2023gpt4v, openai2024gpt4o} and LLaVA-Video \cite{zhang2024video}.
While these models prioritize accuracy, they often incur high computational costs, limiting their practicality for resource-constrained applications.
To further assess its capabilities in real-time video interaction scenarios, we compare it with token-efficient methods \cite{maaz2023video, li2024llama, zhang2025llava} and streaming video-understanding models \cite{zhang2024flash, di2025streaming}.
This analysis extends beyond accuracy to quantify computational overhead (e.g., serving FPS and latency) and memory consumption, emphasizing VideoScan’s suitability for deployment in dynamic, real-world environments.

\noindent\textbf{Evaluation Benchmarks.}
We conduct experiments on several video-based benchmarks, including offline and online settings.
As for evaluating the basic video understanding ability, we assess VideoScan on 4 wildly-used benchmarks, e.g., MVBench \cite{li2024mvbench} focusing in relatively short-term videos, MLVU \cite{zhou2024mlvu} and LongVideoBench \cite{wu2024longvideobench}, characterizing in long-term video understanding tasks, and VideoMME \cite{fu2024video} having diverse videos in various domian with different durations.
As for evaluating the performance in real-time video interactions, we conduct a comparison on VStream-QA benchmark \cite{zhang2024flash}, a question-answering benchmark specifically designed for online streaming video understanding. 

\subsection{Main Results}
\noindent\textbf{Results on offline video understanding benchmarks.}
In \cref{tab: offline}, we present a comprehensive comparative evaluation of VideoScan against state-of-the-art methods across four video understanding benchmarks: MVBench \cite{li2024mvbench}, MLVU \cite{zhou2024mlvu}, LongVideoBench \cite{wu2024longvideobench}, and VideoMME \cite{fu2024video}. 
Notably, VideoScan achieves these results using only one semantic carrier token per frame alongside a memory mechanism with a fixed capacity of $M=128$, which still retains critical tokens for long-term reasoning.
Remarkably, VideoScan achieves competitive performance with advanced video-based VLMs despite its extreme token efficiency. Quantitatively, VideoScan retains over $85\%$ of LLaVA-Video's accuracy while reducing inference overhead by $99\%$. 
Specifically, with VideoScan, LLaVA-Video can achieve up to $6$ FPS rate in serving, which is $\sim 5\times$ the original inference speed of LLaVA-Video in video dialogue scenario.
VideoScan also outperforms specialized long-term video understanding baselines such as LongVA \cite{zhang2024long} and MovieChat \cite{song2024moviechat}.
To further validate efficiency, we compare VideoScan with token-efficient methods on the MLVU benchmark. As detailed in \cref{tab: offline mlvu}, VideoScan surpasses all token-efficient counterparts, achieving a $5\%$ accuracy gain without instruction leakage.
Critically, VideoScan requires no additional learnable parameters for token generation or selection. 
Instead, it harnesses the inherent semantic summarization capabilities of its backbone LLM to compress visual content into compact semantic carriers. 
Overall, VideoScan demonstrates robust video understanding capabilities with one token per frame, showing promise for real-time streaming video tasks.

\noindent\textbf{Results on online streaming video understanding benchmark.}
To validate the effectiveness and efficiency of VideoScan in real-time scenarios, we benchmark it against state-of-the-art streaming methods on the VStream-QA dataset \cite{zhang2024flash}, evaluating both inference accuracy and response efficiency.
Results are summarized in \cref{tab: online results}. Beyond accuracy, we report end-to-end latency and peak GPU memory (VRAM) usage, measured under identical conditions on an A100 GPU without engineering optimizations.
VideoScan achieves superior accuracy on VStream-QA, outperforming Flash-VStream-7B \cite{zhang2024flash} and matching the performance of LLaVA-OV-7B \cite{di2025streaming} augmented with ReKV’s internal retrieval.
Crucially, VideoScan attains this with 1.29$\times$ lower latency and $50\%$ less VRAM consumption than the baseline.
In terms of latency, VideoScan maintains a similar end-to-end response time while being $1.29\times$ faster than ReKV. 
The fixed memory bank size, $M=128$, ensures stable GPU memory usage, even for prolonged interactions, by dynamically discarding redundant tokens while retaining critical spatiotemporal context.
These results underscore the capability of VideoScan to balance accuracy and efficiency in streaming settings through its token-light architecture and fixed-capacity memory bank.

\subsection{Ablation Study}
The semantic carrier token integrates input embeddings derived from average pooling and preserves semantic flow within the KV pairs by leveraging visual information from removed frames.

\noindent\textbf{The effect of semantic carrier design.}
To validate the design choices of VideoScan, we first evaluate the impact of using a generic input embedding (i.e., selecting the last token from each frame’s video tokens during inference) while maintaining the token reduction rate in the prefilling and decoding stages. 
We also evaluate the effect of KV through inference without any KV reuse.
As shown in \cref{tab: ablation semantic}, removing any component of our design leads to significant accuracy degradation, thereby demonstrating the effectiveness of our proposed architecture.

\noindent\textbf{The effect of memory mechanism.}
We also evaluate the effect of the proposed memory mechanism.
We conduct evaluation without memory (w/o mem) settings, where VideoScan takes frame inputs evenly sampled from the video.
We set different upper bounds to investigate the impact of the counts of frames on the performance.
Then we compare them with the full-version VideoScan with two different memory bank sizes.
The results, in \cref{tab: ablation mem}, demonstrate that the memory mechanism maintains significant temporal information, helping it to deal with long-term videos while supporting the real-time video interaction.

\begin{table}[]
	\centering
	\caption{The effect of components in semantic carrier. Performance of VideScan w/o. subtitles (\%) under different semantic carrier settings.}
	\label{tab: ablation semantic}
	\begin{tabular}{c|cccc}
		\toprule
		\textbf{Method} &\textbf{Overall} & \textbf{Short} & \textbf{Medium} & \textbf{Long} \\ 
		\midrule
		w/o emb & 44.5 & 47.7 & 45.1 & 40.7\\
		w/o KV & 42.6 & 44.0 & 43.0 & 40.9\\
		\midrule
		VideoScan & 54.0 & 62.1 & 53.2 &46.6 \\
		\bottomrule
	\end{tabular}
\end{table}

\begin{table}[]
	\centering
	\caption{The effect of the memory mechanism. Performance of VideScan w/o. subtitles (\%) with or without memory bank. Without a memory bank, Videocan evenly sampled frames from offline videos under different upper bounds.}
	\label{tab: ablation mem}
	\small
	\begin{tabular}{c|c|cccc}
		\toprule
		\textbf{Method} &\textbf{\#Frames} & \textbf{Overall} & \textbf{Short} & \textbf{Medium} & \textbf{Long} \\ 
		\midrule
		\multirow{6}{*}{\textbf{w/o. mem}} & 32 & 51.4 & 59.1 & 50.2 & 44.9 \\
		~ & 40 & 51.6 & 59.9 & 50.7 & 44.1 \\
		~ & 48 & 51.8 & 58.8 & 51.1 & 45.6 \\
		~ & 56 & 52.2 & 59.3 & 52.0 & 45.3\\
		~ & 64 & 52.8 & 60 & 53.2 & 45.1 \\
		~ & 128 & 53.8& 61.9 & 53.7 & 45.8 \\
		\midrule
		\multirow{4}{*}{\textbf{w. mem}} & 1fps&\multirow{2}{*}{54.0}&\multirow{2}{*}{62.1}&\multirow{2}{*}{53.2}&\multirow{2}{*}{46.6}\\
		~ & (M=64)\\
		~ & 1fps&\multirow{2}{*}{55.1}&\multirow{2}{*}{64.2}&\multirow{2}{*}{54.3}&\multirow{2}{*}{46.7}\\
		~ & (M=128)\\
		\bottomrule
	\end{tabular}
\end{table}

\section{Conclusion}
In this work, we present VideoScan, a novel framework for efficient streaming video understanding that processes each frame using a single semantic carrier token without relying on predefined instructions.
By leveraging in-context summarization, VideoScan compresses visual data into compact tokens via (1) average-pooled frame features and (2) contextual semantics from past key-value states, drastically reducing computational and memory overhead.
A dynamic memory mechanism retains semantically distinct tokens while discarding redundant ones, ensuring stable performance with a fixed-capacity memory bank. 
Enhanced by a two-stage training strategy to strengthen temporal-semantic coherence, VideoScan achieves 6 FPS ($5\times$ faster than LLaVA-Video) with stable and duration-independent GPU memory usage, enabling real-time applications in robotics, surveillance, and interactive systems.

\noindent\textbf{Limitations.}
One limitation of VideoScan is its incompatibility with Transformer encoder architectures, as its attention sink-based mechanism and causal mask-based semantic training inherently rely on autoregressive decoding paradigms.
Additionally, the extreme vision token reduction rate (one token per frame) sacrifices fine-grained spatial-temporal details, potentially limiting performance on tasks requiring precise visual reasoning.
Possible improvement could explore preserving select information-rich vision tokens (e.g., those with minimal embedding similarity to the semantic carrier) alongside the compressed representations, coupled with instruction-guided memory augmentation to dynamically retrieve details during inference.
However, this risks disrupting the optimized GPU memory stability and computational efficiency that VideoScan currently achieves, necessitating careful design of hybrid memory management and adaptive retrieval mechanisms to balance granularity and performance in streaming scenarios.
{
    \small
    \bibliographystyle{ieeenat_fullname}
    \bibliography{main}
}
\end{document}